\documentclass{article} %
\usepackage[final]{colm2026_conference}
\usepackage[T1]{fontenc}

\usepackage{microtype}
\usepackage{hyperref}
\usepackage{url}
\usepackage{booktabs}

\usepackage[inline]{enumitem}
\usepackage{graphicx}
\usepackage{nicefrac}
\usepackage[table]{xcolor} %
\usepackage{array}         %
\usepackage{amsmath}
\usepackage{amssymb}
\usepackage{cleveref}
\usepackage{dsfont}
\usepackage{subcaption}
\usepackage{multirow}
\usepackage{xurl}
\usepackage{tabularx}
\usepackage{float}

\definecolor{darkblue}{rgb}{0, 0, 0.5}
\hypersetup{
    colorlinks=true,
    citecolor=darkblue,
    linkcolor=darkblue,
    urlcolor=darkblue,
    pdftitle={Apriel-Reasoner: RL Post-Training for General-Purpose and Efficient Reasoning},
    pdfauthor={Rafael Pardinas, Ehsan Kamalloo, David Vazquez, Alexandre Drouin}
}

\title{{\apriel}: RL Post-Training for General-Purpose and Efficient Reasoning}

\author{Rafael Pardinas\thanks{Now at another institution. Correspondence: \texttt{rafapi@gmail.com}.} \quad Ehsan Kamalloo \quad David Vazquez \quad Alexandre Drouin\\
ServiceNow AI Research}

\usepackage{listings}
\definecolor{promptblue}{HTML}{2B4C7E}

\newcommand{\includeprompt}[3][]{%
  \lstinputlisting[
    title={#2},
    basicstyle=\ttfamily\small,
    breaklines=true,
    breakatwhitespace=false,
    columns=fullflexible,
    lineskip=-0.2pt,
    keepspaces=true,
    showstringspaces=false,
    aboveskip=0.25\baselineskip,
    belowskip=0pt,
    frame=single,
    rulecolor=\color{promptblue},
    backgroundcolor=\color{promptblue!5},
    #1
  ]{#3}%
}

\newcommand{\apriel}{Apriel-Reasoner}
\newcommand{\aprielbase}{Apriel-Base}

\begin{document}

\maketitle

\begin{abstract}
Building general-purpose reasoning models using reinforcement learning with verifiable rewards (RLVR) across diverse domains has been widely adopted by frontier open-weight models. However, their training recipes and domain mixtures are often not disclosed. Joint optimization across domains poses significant challenges: domains vary widely in rollout length, problem difficulty, and sample efficiency. Further, models with long chain-of-thought traces increase inference cost and latency, making efficiency critical for practical deployment. We present {\apriel}, trained with a reproducible multi-domain RL post-training recipe on {\aprielbase}, a 15B-parameter open-weight LLM, across five domains using public datasets: mathematics, code generation, instruction following, logical puzzles, and function calling. We introduce adaptive domain sampling that preserves target completed-rollout ratios despite heterogeneous rollout dynamics, and a difficulty-aware length penalty that, at no additional training overhead, encourages longer reasoning for difficult problems and shorter traces for easy ones. Trained with a strict 16K-token output budget, {\apriel} remains effective at a 32K output budget and improves over {\aprielbase} on AIME 2025, GPQA, MMLU-Pro, and LiveCodeBench while producing 30--50\% shorter reasoning traces. Among the evaluated open-weight models of similar scale, it improves the accuracy--token trade-off using a fully public-data recipe.
\end{abstract}

\begin{figure}[h]
    \centering
    \includegraphics[width=0.84\linewidth]{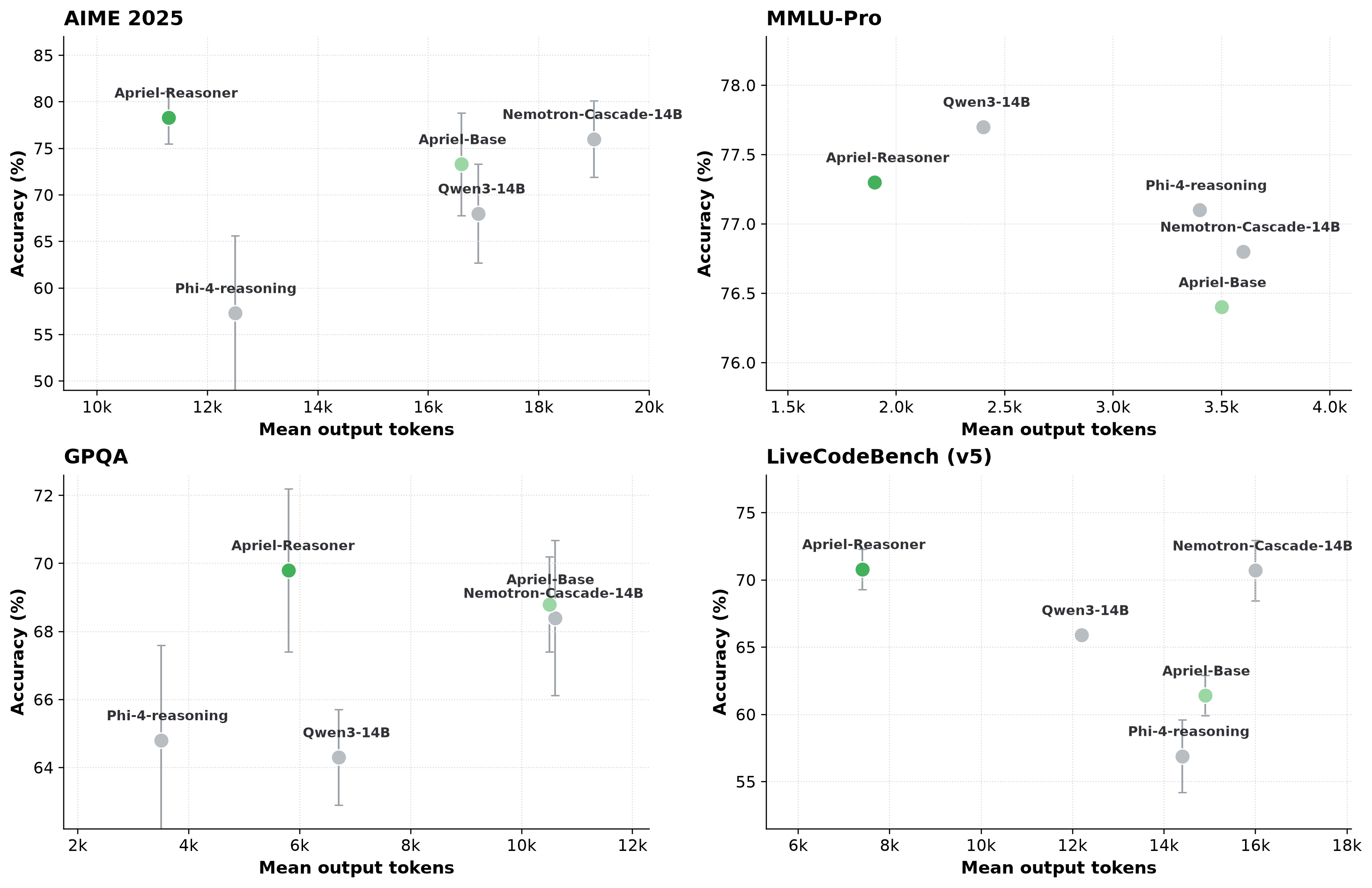}
    \caption{Accuracy versus mean output tokens under a shared 32K output budget. Error bars show standard deviations across repeated sampling-seed evaluations where available.}
    \label{fig:pipeline-evals}
\end{figure}

\section{Introduction}
\label{sec:introduction}

Reinforcement Learning with Verifiable Rewards (RLVR)~\citep{lambert2024tulu, deepseek, kimi2, olmo3} has become an integral stage in post-training large language models (LLMs) for reasoning. Recent open-weight models increasingly rely on multi-domain RLVR to build general-purpose reasoning capabilities, including Qwen-3.5~\citep{qwen3.5}, GLM~\citep{zeng2025glm, zeng2026glm5}, Kimi k2~\citep{kimi2}, OLMo 3~\citep{olmo3}, and Nemotron-Cascade~\citep{wang2025nemotron}. However, despite their open weights, the underlying training recipes, domain mixtures, and multi-domain optimization details are often not fully disclosed. This limits reproducibility and understanding of which ingredients drive strong multi-domain reasoning performance.

Beyond reproducibility concerns, multi-domain RLVR also introduces concrete optimization challenges. 
When training across multiple domains simultaneously, rollouts are generated asynchronously, and domains vary widely in rollout length, problem difficulty, verification latency, and sample efficiency. As a result, faster or easier domains tend to complete rollouts sooner and produce a disproportionate share of collected trajectories, shifting the completed-rollout mixture away from the intended domain ratios. Without correction, this drift biases the data entering downstream preprocessing toward overrepresented domains. Existing approaches either rely on static domain mixtures~\citep{olmo3} or avoid joint optimization entirely by training domains sequentially~\citep{wang2025nemotron}. We address this with adaptive domain sampling, which dynamically preserves target completed-rollout ratios despite heterogeneous rollout dynamics.

Orthogonal to domain balancing, reasoning models trained with RLVR tend to generate long chain-of-thought traces~\citep{chen2025not}, incurring high inference cost and latency. Prior approaches to controlling reasoning length range from fixed budgets that ignore problem difficulty~\citep{s1, l1, team2025kimi} to more elaborate difficulty-aware mechanisms~\citep{thinkless, adaptthink, zhang-zuo-2025-grpo, dast, dler}. We propose a simpler alternative with no additional training overhead: a difficulty-aware extension of the standard length penalty that relaxes the length penalty on harder problems while preserving the standard penalty on easier ones, without changing the policy loss.

To study these ideas in practice, we present {\apriel}, a model trained with a reproducible multi-domain RL post-training recipe on {\aprielbase}~\citep{radhakrishna2025apriel}, a 15B-parameter open-weight LLM that has not been previously trained with RL, making it a clean starting point for studying multi-domain RLVR at a practical model scale. We build on PipelineRL~\citep{piche2025pipelinerl}, which enables asynchronous on-policy training via in-flight weight updates, so rollout generation and optimization proceed concurrently during multi-domain RL. We train across five public datasets spanning mathematics, code generation, instruction following, logical puzzles, and function calling.

Despite RL post-training under a strict 16K-token output budget, {\apriel} remains effective at a 32K output budget. Because {\aprielbase} already supports long-context inference, this result demonstrates preservation of effective behavior under the larger evaluation budget rather than RL-induced length extrapolation. {\apriel} consistently improves over {\aprielbase} on AIME 2025, GPQA, MMLU-Pro, and LiveCodeBench while producing 30--50\% shorter outputs. Among the evaluated open-weight baselines of similar scale, it improves the accuracy--token trade-off using a reproducible public-data recipe, as illustrated in Figure~\ref{fig:pipeline-evals}. We detail all training configurations and domain mixtures to support reproducibility in multi-domain RL research.

Below, we summarize our key contributions:
\begin{enumerate}[leftmargin=*, itemsep=2pt, topsep=2pt]
    \item \textbf{Reproducible multi-domain RL post-training:}  We present a fully open and reproducible RL post-training recipe across five diverse domains that achieves competitive accuracy with substantially shorter reasoning traces than comparable-scale models.
    \item \textbf{Adaptive domain sampling:} We introduce an adaptive sampler that maintains target completed-rollout domain ratios during asynchronous multi-domain RL training.
    \item \textbf{Difficulty-Aware Length Penalty:} We introduce a simple extension of the standard length penalty that accounts for problem difficulty, allowing difficult problems to reason longer while encouraging conciseness on easier ones.
\end{enumerate}

\begin{figure}[h]
    \centering
    \includegraphics[width=\linewidth]{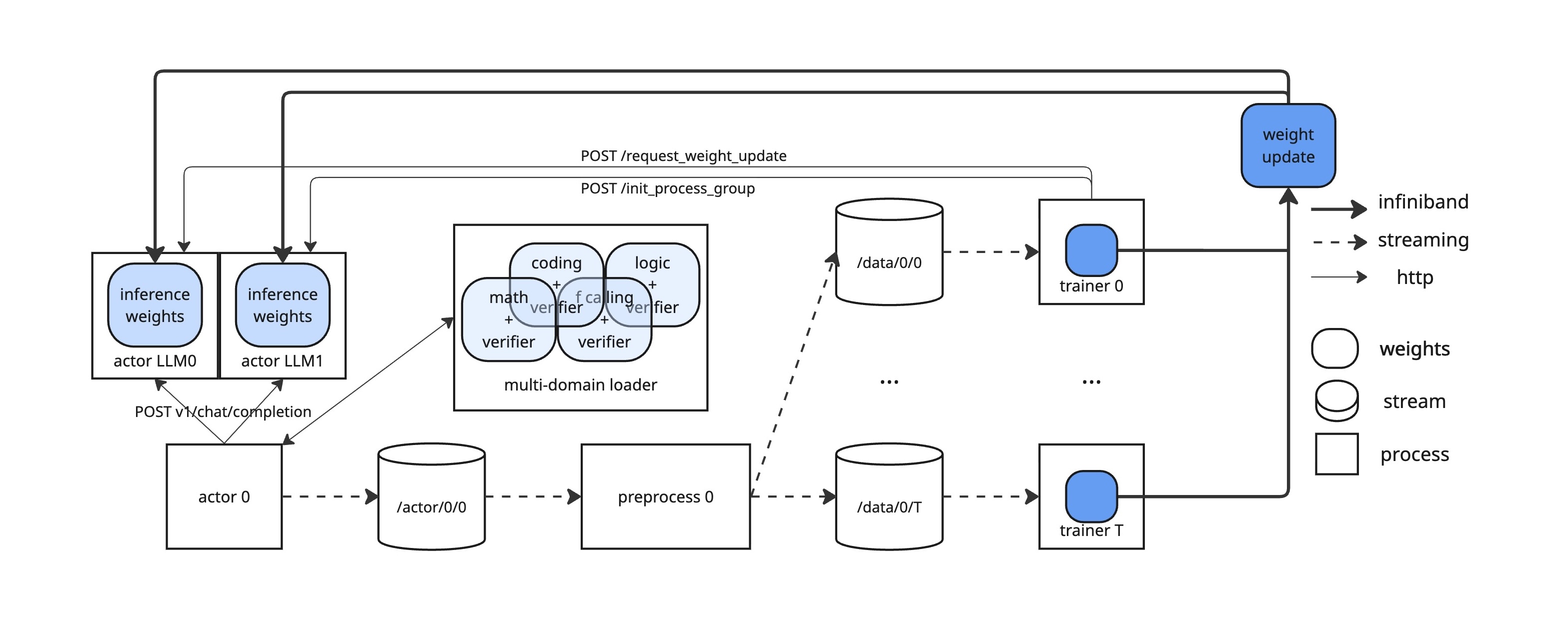}
    \caption{PipelineRL with the multi-domain extensions used in this work. A domain-weighted sampler draws prompts from the five training environments, environment-specific verifiers score the resulting rollouts, and the data flow through the actor, preprocessor, and trainer stages for asynchronous RL post-training.}
    \label{fig:pipeline-overview}
\end{figure}

\section{Background}
\label{sec:background}

\subsection{Reinforcement Learning with Verifiable Rewards}
Reinforcement learning is widely used as a post-training step to equip LLMs with long-form reasoning capabilities for solving complex problems \citep{shao2024deepseekmath, lambert2024tulu, deepseek}. The core idea is to frame auto-regressive language modeling as a sequential decision-making process, where a programmatically verifiable reward, computed by checking the correctness of the model's output, serves as the training signal to guide the model toward producing correct solutions. This paradigm is commonly referred to as RL with Verifiable Rewards (RLVR).

The training objective in RLVR is to maximize the expected reward of the  response $y$, generated by the language model policy $\pi_\theta$, given a prompt $x$:
\begin{equation}
    \mathcal{J}(\theta) = \mathbb{E}_{x \sim \mathcal{D}, y \sim \pi_\theta(\cdot|x)} \left[ R(x, y) \right],
\end{equation}
where $R(x, y)$ is a reward function evaluating response correctness.

\subsection{Policy Optimization}
\label{sec:gspo}
In the LLM era, Proximal Policy Optimization (PPO)~\citep{schulman2017proximal} has been widely adopted over REINFORCE~\citep{williams1992simple}. PPO stabilizes training by approximating a trust region constraint~\citep{trpo} and using a learned critic for per-token advantage estimation. However, in the RLVR setting, training an accurate critic from sparse, end-of-sequence rewards is difficult~\citep{shao2024deepseekmath, kazemnejad2025vineppo} and introduces significant overhead~\citep{ahmadian-etal-2024-back}. Group Relative Policy Optimization (GRPO)~\citep{shao2024deepseekmath} sidesteps the critic by sampling multiple responses per prompt and computing advantages from their relative rewards. Yet both PPO and GRPO operate at the token level while rewards are sequence-level, a mismatch that introduces high-variance gradients and training instability~\citep{yu2025dapo, zheng2025gspo, zhao2026geometricmean}.

\textbf{Group Sequence Policy Optimization (GSPO)~\citep{zheng2025gspo}} replaces the token-level importance ratio with a \emph{sequence-level} importance ratio, aligning the unit of optimization with the unit of reward. For each prompt $x$, GSPO samples a group of $G$ responses $\{y_i\}_{i=1}^G$ from the current policy $\pi_{\theta_{\text{old}}}$ and computes the advantage $\hat{A}_i$ of each response by normalizing $R(x, y_i)$ relative to the group rewards. The objective is defined as

\begin{equation}
    \mathcal{J}_{\text{GSPO}}(\theta) = \mathbb{E}_{x \sim \mathcal{D},\, \{y_i\} \sim \pi_{\theta_{\text{old}}}} \left[ \frac{1}{G} \sum_{i=1}^{G} \min\!\left( s_i(\theta)\, \hat{A}_i,\; \mathrm{clip}\!\left(s_i(\theta),\, 1{-}\varepsilon,\, 1{+}\varepsilon\right) \hat{A}_i \right) \right],
\end{equation}
where the sequence-level importance ratio is defined as:
\begin{equation}
    s_i(\theta) = \exp\!\left(\frac{1}{|y_i|} \sum_{t=1}^{|y_i|} \log \frac{\pi_\theta(y_{i,t} \mid x, y_{i,<t})}{\pi_{\theta_{\text{old}}}(y_{i,t} \mid x, y_{i,<t})}\right).
\end{equation}
The length normalization by $\frac{1}{|y_i|}$ reduces variance and unifies the numerical range of $s_i(\theta)$ across responses of different lengths.
By applying a single importance weight to all tokens in a response, GSPO eliminates the high-variance per-token reweighting of GRPO and provides a more stable learning signal.
GSPO has been adopted in the training of recent models such as Qwen3-Omni~\citep{qwen3omni}. In this work, we adopt GSPO for the RL post-training of {\apriel}.

\subsection{On-Policy RL for LLMs}
The effectiveness of LLM post-training via RL hinges on on-policy data, where the model trains on rollouts sampled from its current parameters rather than from a stale or separate behavior policy. Because the output distribution of an autoregressive model is highly sensitive to its parameters, training on off-policy data introduces a distribution mismatch that can degrade learning~\citep{tajwar2024preference, zheng2025prosperity, ritter2026llms}.

\textbf{PipelineRL~\citep{piche2025pipelinerl}} is a distributed RL system for LLM training that addresses the throughput-freshness trade-off via \textit{in-flight weight updates}. Rather than enforcing a synchronization barrier between rollout and training, or fully decoupling them at the cost of policy staleness, PipelineRL runs both concurrently, i.e., inference workers continuously receive updated weights without pausing generation, keeping data near on-policy while maintaining high GPU utilization. To achieve high throughput with variable-length generations, PipelineRL incorporates standard practices from large-scale LLM training and serving, including sequence packing, dynamic batching, and sequence parallelism.
We use PipelineRL for RL post-training throughout this work, and discuss additional changes made to the system toward post-training a general-purpose reasoning LLM.

\section{Related Work}
\label{sec:relatedwork}
\paragraph{RL post-training of LLMs:} Open-weight RL post-training recipes for general-purpose reasoning typically follow either staged domain-wise optimization or joint multi-domain optimization. Staged approaches optimize capabilities sequentially, as in Nemotron-Cascade~\citep{wang2025nemotron} and GLM~\citep{zeng2025glm, zeng2026glm5}, which simplifies optimization but introduces order effects and requires careful management of catastrophic forgetting~\citep{kirkpatrick2017overcoming}. Joint optimization instead trains across domains simultaneously, as in OLMo 3~\citep{olmo3} and Kimi K2~\citep{kimi2}, allowing shared reasoning patterns to be reinforced throughout training. We follow the joint setting and focus on one of its practical challenges: preserving the intended training mixture under asynchronous multi-domain rollout dynamics. Sequential-domain training is therefore a useful but distinct comparison rather than a substitute for evaluating completion skew in the concurrent setting.

\paragraph{Reasoning Length Control:} Reasoning models are prone to \emph{overthinking}, generating excessively long reasoning traces even on relatively simple problems~\citep{chen2025not}. Existing approaches control length through test-time budget forcing~\citep{s1}, budget-conditioned training~\citep{l1}, training-time penalties~\citep{team2025kimi}, or explicit budget tokens during rollout generation~\citep{wen2025budgetthinker}. More recent methods introduce difficulty awareness through routing~\citep{thinkless, adaptthink}, adaptive policy objectives~\citep{zhang-zuo-2025-grpo}, preference optimization over difficulty-ranked traces~\citep{dast}, truncation-aware RL adjustments~\citep{dler}, or explicit budget prediction~\citep{han-etal-2025-token,li2025selfbudgeter}. 
Our approach formulates length control as a difficulty-aware penalty integrated directly into the reward function, requiring no auxiliary models, no additional training phases, and no changes to the policy loss. \Cref{tab:length-control-comparison} makes the distinction from the closest difficulty-aware methods explicit. DAP's simplicity is an integration and compute claim, not an accuracy-superiority claim: unlike GRPO-LEAD, DLER/DA-DLER, DAST, and AdaptThink, it leaves rollout collection and the policy objective unchanged. We do not provide matched-compute experiments against these methods.

\section{Multi-Domain RL}
\label{sec:domains}

In this section, we describe the multi-domain training recipe used for {\apriel} and the two mechanisms we introduce to stabilize domain balance and control reasoning length.

\subsection{Adaptive Multi-Domain Sampling}
\label{sec:adaptive-sampling}
Training across multiple domains requires coordinating rollout generation across environments with heterogeneous characteristics. The time required to generate a rollout and compute its reward can vary substantially across domains. For example, coding problems often require longer generations, and evaluating them involves executing code in a sandbox, whereas math or logic answers can often be verified with lightweight checks.
Within a fixed training window, some domains therefore produce completed rollouts more quickly than others, shifting the completed-rollout mixture away from the configured domain ratios. Zero-advantage group filtering~\citep{yu2025dapo} can introduce an additional downstream imbalance because harder domains tend to yield fewer retained trajectories when successful samples are less frequent. The adaptive sampler described below corrects completed-rollout skew; it does not directly guarantee the post-filter retained mixture. Without adaptive correction, preserving a target completed-rollout mixture for each chosen domain-rate configuration would require manually resizing or sub-sampling the per-domain data for that run, undermining one of the main practical advantages of a shared multi-environment training system.

To maintain target completed-rollout domain ratios throughout training, we implement a simple adaptive sampling strategy in PipelineRL, as illustrated in \Cref{fig:pipeline-overview}. 
Let $\mathcal{D}$ be the set of domains, and let $w_d > 0$ denote the configured target weight for domain $d$, where $\sum_{d \in \mathcal{D}} w_d = 1$. When the actor samples a new problem for a rollout, let $n_d$ be the current number of completed rollouts from domain $d$ and $N = \sum_{j \in \mathcal{D}} n_j$ the total up to this point. To correct for drift from the configured domain mixture, we first compute an adjustment factor:
\begin{equation}
    \alpha_d = \mathrm{clip}\!\left(\frac{w_d}{n_d / N},\, 0.1,\, 10.0\right)
\end{equation}
Domains that have produced fewer completions than intended get $\alpha_d > 1$ and are up-weighted, while overrepresented domains are down-weighted. If a domain has no completions yet, we set $\alpha_d = 10.0$. The clipping bounds prevent extreme corrections when a domain is heavily over- or under-represented. We then sample a domain with probability:
\begin{equation}
    p_d = \frac{w_d \, \alpha_d}{\sum_{j \in \mathcal{D}} w_j \, \alpha_j}
\end{equation}
At the initial stages of training, we use the static distribution $p_d = w_d$, i.e., the configured domain ratios, until at least 50 total completions have been collected. Once a domain is selected, a sample is drawn uniformly from that domain.

\paragraph{Realized domain proportions.}
\Cref{fig:adaptive-sampling-trajectory} shows the full logged trajectory of cumulative completed-rollout proportions for the selected DAP run, while \Cref{tab:adaptive-sampling-realized} reports the selected checkpoint numerically. The first non-empty logged window (step 0; 2,928 completed rollouts) begins under static sampling weights and exhibits substantial skew: function calling contributes 20.5\% of completions against a 10.0\% target, while logic contributes 5.7\% against a 15.0\% target. Because adaptive correction starts after 50 completions, this aggregated window may already include early corrective sampling and is not a purely pre-correction baseline. By the selected checkpoint (step 250; 683K cumulative rollouts), the mean absolute deviation from the target across domains is 0.027 percentage points. At the end of the logged trajectory (step 331; 952K rollouts), it is 0.035 percentage points. These diagnostics show that the observed completed-rollout mixture converges to and tracks the configured target under adaptive sampling; they do not replace a matched static-sampler training ablation measuring downstream accuracy.

\subsection{Difficulty-Aware Length Penalty}
\label{sec:dap}
In this section, we present our extension to the standard length penalty in which the penalty strength is modulated by problem difficulty, which we call \emph{Difficulty-Aware Length Penalty}.

\paragraph{Length Penalty (LP):}
Let $L$ denote the maximum output length, and $B$ be the buffer width. For a prompt $x$ and rollout $\hat{y}_i$ of length $l_i$, the penalty is applied linearly within a buffer zone before the maximum length:
\begin{equation}
P(l_i)=
\begin{cases}
\frac{(L-B)-l_i}{B}, & L-B < l_i \le L\\
0, & \text{otherwise}
\end{cases}
\end{equation}
Rollouts shorter than $L-B$ incur no penalty, while those approaching $L$ are penalized proportionally. The final reward combines the task reward $R(x, \hat{y}_i)$ with this penalty:
\begin{equation}
r_i = R(x, \hat{y}_i) + \lambda \, P(l_i),
\end{equation}
where $\lambda$ controls the penalty strength. In the standard setting, $\lambda = 1$ for all rollouts.

\paragraph{Difficulty-Aware Length Penalty (DAP):}
The key idea behind DAP is to modulate $\lambda$ based on problem difficulty, so that difficult problems receive a weaker penalty and are allowed to reason longer. We estimate difficulty from the solve rate within each group: given $N$ rollouts for a prompt $x$, we compute:
\begin{equation}
s = \frac{1}{N}\sum_{i=1}^N c_i,
\end{equation}
where $c_i = \mathds{1}[R(x, \hat{y}_i) = 1]$ indicates whether rollout $i$ produced a correct answer. A low solve rate indicates a harder problem. We then set:
\begin{equation}
\lambda=
\begin{cases}
1, & l_i \ge L \;\wedge\; \neg\, \text{finished}_i \quad \text{(upper-bound guard)}\\
s^\gamma, & c_i=1\\
\lambda_f, & c_i=0
\end{cases}
\end{equation}

For correct rollouts, the penalty is scaled by $s^\gamma$ where $\gamma \ge 0$ is a constant: when the solve rate $s$ is low (hard problem), $\lambda$ becomes small and the penalty is relaxed, allowing the model to use more reasoning tokens. For incorrect rollouts, the penalty reduces to the standard length penalty with a fixed scaling factor $\lambda_f \in [0,1]$. The upper-bound guard ensures that rollouts which are truncated at the maximum length without finishing always receive the full penalty, preventing degenerate solutions that exhaust the budget. DAP modifies only the reward $r_i$ upstream and does not alter the policy loss.

In our experiments, we use $\gamma=1.0$ and $\lambda_f=1.0$, i.e., incorrect rollouts always receive the full penalty, while the penalty on correct but overlong rollouts is scaled by the solve rate.

\begin{table}[t]
\centering
{
\small
\begin{tabular}{>{\raggedright\arraybackslash}p{0.19\linewidth}>{\raggedright\arraybackslash}p{0.36\linewidth}>{\raggedright\arraybackslash}p{0.35\linewidth}}
\toprule
\textbf{Domain} & \textbf{Training data} & \textbf{Verifiable reward} \\
\midrule
Mathematics & Open-Reasoner-Zero ($\sim$129K) & Exact-match verification of the extracted final answer \\
Code Generation & TACO ($\sim$24K) & Pass/fail reward from sandboxed execution against test cases \\
Instruction-Following & IF-RLVR ($\sim$95K) & Fraction of output constraints satisfied \\
Logical Puzzles & INTELLECT-3 / SynLogic ($\sim$12K) & Task-specific programmatic verifier \\
Function Calling & BFCL v4 single-turn tasks ($\sim$4K) & Exact function-name match and argument-set validity \\
\bottomrule
\end{tabular}
}
\caption{Summary of the five training environments used for joint multi-domain RL.}
\label{tab:domain-summary}
\end{table}

\section{Experiments}
\label{sec:experiments}

\subsection{Experimental Setup}
\label{sec:exp-setup}

\paragraph{Training Domains.}
\label{sec:envs}
We train across five environments, each defined by a public dataset and a verifiable reward function. \Cref{tab:domain-summary} summarizes the domains used for joint RL post-training; full dataset and verifier details are provided in Appendix~\ref{sec:appendix-envs}.

\paragraph{Model.} We use Apriel-1.5\footnote{\url{https://huggingface.co/ServiceNow-AI/Apriel-1.5-15b-Thinker}} \citep{radhakrishna2025apriel} as our base model, referred to as {\aprielbase} throughout the paper. Apriel-1.5 is a 15B-parameter multimodal model initialized from Pixtral-12B~\citep{pixtral} via depth upscaling, then continually pre-trained and instruction-tuned
with explicit reasoning traces.
We omit the vision encoder and use only the decoder, as multimodal reasoning is outside the scope of this work.
Crucially, Apriel-1.5 is trained without RL or preference optimization. We choose it over the newer Apriel-1.6~\citep{madhusudhan2025apriel16}, which incorporates RL in its post-training, and over other popular open-weight models like Qwen3~\citep{yang2025qwen3}, which release weights but only describe their training data at a high level. In contrast, the mid-training domains, data composition, and training recipe of Apriel-1.5 are fully documented~\citep{radhakrishna2025apriel}, making it a suitable starting point for studying the effects of RL post-training, as downstream behavior changes can be more confidently attributed to our RLVR training.
\paragraph{Training Setup.} All {\apriel} jobs were run on 8 compute nodes with 8 Nvidia H100 GPUs per node (64 GPUs total). 
Within each node, 4 GPUs are allocated to rollout generation (actor) and 4 to training. We train with GSPO and DAP (length penalty), adopting the clip-higher asymmetric clipping and dynamic sampling from DAPO~\citep{yu2025dapo}.
Following~\citet{khatri2025art}, the final vocabulary projection is computed in FP32 on both actors and training workers for numerical stability, while the rest of the model remains in BF16. We find the best domain mixture empirically by training on various mixtures and selecting the best one on fixed validation sets randomly held out from each training source before RL optimization. These sets comprise 10\% of the source data, are never resampled, and are not used for policy optimization. The selected mixture assigns 40\% to mathematics (M), 25\% to code (C), 15\% to logical puzzles (L), 10\% to instruction-following (I), and 10\% to function calling (F).
The selected model is the checkpoint at step 250 ($\sim$36 wall-clock hours). We set a maximum of 400 optimization steps and evaluate available checkpoints on held-out validation sets every 50 steps starting at step 150; step 250 yields the best mean validation accuracy across the five domains.
The full training configuration and hyperparameter sweep ranges are summarized in Tables~\ref{tab:gspo-dap-launch-config} and \ref{tab:gspo-dap-sweep-values} in Appendix~\ref{sec:training-details}.

\paragraph{Algorithm portability.}
The experiments in this paper use GSPO, but neither proposed component depends on its sequence-level objective. Adaptive sampling acts before optimization on domain-typed prompts and completed-rollout counts. DAP reshapes a scalar rollout reward using grouped solve rates and can therefore be used with grouped on-policy scalar-reward methods such as GRPO or PPO-style RLVR when equivalent solve-rate estimates are available. DPO is different: adapting DAP to an offline preference objective would require constructing preferences or weights from difficulty and length signals. These are design-level compatibility arguments; effectiveness beyond GSPO is not empirically established here.

\paragraph{Baselines.} We select open-weight models of comparable size to {\apriel}: (i) Nemotron-Cascade\footnote{\url{https://hf.co/nvidia/Nemotron-Cascade-14B-Thinking}} \citep{wang2025nemotron} is a 14B reasoning model trained via sequential domain-wise RL. (ii) Qwen3-14B\footnote{\url{https://hf.co/Qwen/Qwen3-14B}} \citep{yang2025qwen3} is a 14B dense model trained with multi-stage post-training combining Supervised Fine Tuning (SFT) and RL across reasoning and general domains. (iii) Phi-4-reasoning\footnote{\url{https://hf.co/microsoft/Phi-4-reasoning}} \citep{abdin2025phi4} is a 14B model fine-tuned from a base model using SFT on o3-mini reasoning traces followed by RLVR. To ensure a fair comparison, we rerun all baselines under analogous output token budget.

\paragraph{Test Benchmarks and Metrics.} We evaluate {\apriel} and the baselines on four external held-out reasoning benchmarks: AIME 2025 (Mathematics; \citealt{balunovic_srimatharena_2025}), GPQA (Graduate-level science; \citealt{rein2024gpqa}), MMLU-Pro (Multi-domain knowledge and reasoning; \citealt{wang2024mmlupro}), and LiveCodeBench (Coding; \citealt{livecodebench}). 
To reduce variance on smaller benchmarks, we repeat each evaluation multiple times under different sampling seeds and average the results: 10 times for AIME 2025 (30 problems), 5 for GPQA (198 problems),
and 3 for LiveCodeBench (315 problems for \texttt{v5} from 2024-07-01 to 2024-12-31); MMLU-Pro (12{,}032 problems) uses a single evaluation.
We report pass@1 accuracy; per-problem accuracy is first averaged across problems, then across runs, and finally across two independent training seeds. For AIME 2025, each Apriel variant uses ten distinct decoding seeds (42--51) for each of two training seeds, giving 20 seed-level accuracies and $30\times20=600$ sampled completions. Standard deviations are computed across the sampling-seed accuracies and are reported where available. To characterize reasoning efficiency under the same generation budget, we also report the average number of output tokens per completion. Following \citet{radhakrishna2025apriel}, all generations use temperature $0.6$ and top-$p$ $0.95$, with a maximum output length of 32K tokens, as used by \citet{deepseek}.

\subsection{Main Results}
\label{sec:main-results}

\begin{table*}[t]
\centering
{
\small
\setlength{\tabcolsep}{3pt}
\begin{tabular}{lccccccccc}
\toprule
 & & \multicolumn{2}{c}{\textbf{AIME-25}} & \multicolumn{2}{c}{\textbf{GPQA}} & \multicolumn{2}{c}{\textbf{MMLU-Pro}} & \multicolumn{2}{c}{\textbf{LCB (v5)}} \\
\cmidrule(lr){3-4}\cmidrule(lr){5-6}\cmidrule(lr){7-8}\cmidrule(lr){9-10}
\multirow{-2}{*}{\textbf{Model}} & \multirow{-2}{*}{\textbf{Size}}  & Acc & Tok & Acc & Tok & Acc & Tok & Acc & Tok \\
\midrule
Phi-4-reasoning & 14B & 57.3 & 12.5k & 64.8 & \textbf{3.5k} & 77.1 & 3.4k & 56.9 & 14.4k \\
Qwen3 & 14B & 68.0 & 16.9k & 64.3 & 6.7k & \textbf{77.7} & 2.4k & 65.9 & 12.2k \\
Nemotron-Cascade & 14B & $76.0\,(4.1)$ & 19.0k & $68.4\,(2.3)$ & 10.6k & 76.8 & 3.6k & $70.7\,(2.3)$ & 16.0k \\
\midrule
{\aprielbase} & 15B & 73.3 & 16.6k & 68.8 & 10.5k & 76.4 & 3.5k & 61.4 & 14.9k \\
\quad + RLVR w/ LP & 15B & 71.7 & \textbf{11.1k} & 68.9 & 4.5k & 75.6 & \textbf{1.5k} & 67.7 & \textbf{7.0k} \\
{\apriel} (\textbf{Ours}) & 15B & \textbf{$78.3\,(2.8)$} & 11.3k & \textbf{$69.8\,(2.4)$} & 5.8k & 77.3 & 1.9k & \textbf{$70.8\,(1.5)$} & 7.4k \\
\bottomrule
\end{tabular}
}
\caption{Test accuracy (\%) and mean output tokens for {\apriel}, its fixed-length-penalty counterpart, {\aprielbase}, and comparable-scale reasoning models. All models are evaluated under the same configuration with a 32K output-token budget. Both RL-trained Apriel variants use identical adaptive domain sampling, base model, training data, and GSPO setup; their comparison isolates LP versus DAP. For repeated evaluations, available standard deviations across sampling-seed accuracies are shown in parentheses. Each Apriel AIME result aggregates ten decoding seeds for each of two training seeds (600 completions); Nemotron-Cascade uses ten decoding attempts.
See~\Cref{fig:pipeline-evals} for the corresponding accuracy-efficiency scatter plots.}
\label{tab:main-results}
\end{table*}

\paragraph{Accuracy--efficiency trade-off.} A central goal of our RL post-training is to improve reasoning quality while avoiding overthinking. \Cref{tab:main-results} shows that {\apriel} achieves competitive point-estimate accuracy with fewer output tokens than the evaluated comparable-scale baselines across the four benchmarks. These comparisons support an improved accuracy--token trade-off within this evaluation suite, not a universal Pareto-frontier claim.
On AIME-25, {\apriel} has the highest point estimate (78.3\%) while generating 
41\% fewer tokens than Nemotron-Cascade. Because AIME contains only 30 problems and the standard deviations overlap (2.8 versus 4.1 percentage points), the 2.3-point AIME gap is not statistically decisive in isolation. On LiveCodeBench, {\apriel} is on par with Nemotron-Cascade while generating less than half as many output tokens (7.4K vs.\ 16.0K). On GPQA, it has the highest point estimate (69.8\%) while using roughly half the tokens of Nemotron-Cascade. On MMLU-Pro, all models cluster within $\sim$1 point of accuracy; {\apriel} generates 1.9K tokens, 21\% fewer than Qwen3 at 2.4K and roughly half as many as the other evaluated models.
Relative to {\aprielbase}, the reduction in output length is largest on GPQA ($\sim$45\%) and MMLU-Pro ($\sim$46\%), where the base model overthinks on straightforward questions, and smallest on AIME-25 ($\sim$32\%), where long traces are more often warranted.

\paragraph{Per-Domain Reward Progress.} To assess learning across all five domains during multi-domain RL training, we plot the mean reward for each domain over the course of training in \Cref{fig:reward_mean}. All domains exhibit an upward trend over the logged run. This observation is consistent with concurrent improvement, although it does not by itself rule out negative transfer relative to separately trained domain-specific controls.

\paragraph{Impact of DAP (vs. standard length penalty).}
The last two rows of \Cref{tab:main-results} compare DAP, described in Section~\ref{sec:dap}, against the standard length penalty. In the standard penalty, the penalty is fixed for all problems, which produces shorter outputs across all four benchmarks, but at the expense of lower test accuracy. DAP recovers this accuracy with a small increase in output length: on AIME-25, accuracy improves by 6.6 percentage points with only 2\% more tokens; on LCB, by 3.1 points with 6\% more tokens; and on GPQA and MMLU-Pro, by 0.9 and 1.7 points with approximately 28\% more tokens.

To test whether this gain primarily reflects aggressive shortening of failed attempts, \Cref{tab:dap-conditioned-lengths} conditions AIME 2025 rollout length on correctness and problem difficulty, using the LP solve rate to define shared bins for both models. On problems that LP solves at least half of the time, DAP shortens correct rollouts by 352 tokens (6\%) in the easiest bin and by 2,872 tokens (19\%) in the medium-easy bin. In the hardest bin, DAP instead allocates more tokens to both correct and incorrect attempts and raises accuracy from 10\% to 22\%. This pattern is consistent with preserving reasoning budget on hard problems while shortening successful reasoning on more solvable problems. The two middle bins contain only one and four problems, so their estimates are illustrative rather than conclusive.

\begin{table*}[t]
\centering
{
\small
\begin{tabular}{lcccccccc}
\toprule
 & \multicolumn{2}{c}{\textbf{AIME-25}} & \multicolumn{2}{c}{\textbf{GPQA}} & \multicolumn{2}{c}{\textbf{MMLU-Pro}} & \multicolumn{2}{c}{\textbf{LCB (v5)}} \\
\cmidrule(lr){2-3}\cmidrule(lr){4-5}\cmidrule(lr){6-7}\cmidrule(lr){8-9}
\multirow{-2}{*}{\textbf{Model}} & Acc & Tok & Acc & Tok & Acc & Tok & Acc & Tok \\
\midrule
Uniform & $75.0\pm5.5$ & 11.9k & $67.7\pm2.1$ & \textbf{5.8k} & 76.1 & \textbf{1.7k} & $70.5\pm0.5$ & \textbf{6.6k} \\
Math \& Code & $75.7\pm5.2$ & 11.7k & $69.1\pm1.7$ & 6.2k & 75.9 & 1.9k & $67.2\pm1.1$ & 7.1k \\
Our mixture & \textbf{\boldmath$78.3\pm2.8$} & \textbf{11.3k} & \textbf{\boldmath$69.8\pm2.4$} & \textbf{5.8k} & \textbf{77.3} & 1.9k & \textbf{\boldmath$70.8\pm1.5$} & 7.4k \\
\bottomrule
\end{tabular}
}
\caption{Domain mixture ablation~--~test accuracy (\%) and mean output tokens of {\apriel} trained on three domain mixtures using the same hyperparameters. Mixtures are reported in M/C/L/I/F order: Uniform = 0.20/0.20/0.20/0.20/0.20, Math \& Code = 0.50/0.50/0/0/0, and Ours  = 0.40/0.25/0.15/0.10/0.10 (matching Table~\ref{tab:main-results}). Accuracy is reported as mean $\pm$ standard deviation across multiple runs, except for MMLU-Pro, which uses a single evaluation.}
\label{tab:domain-mixture-ablation}
\end{table*}

\paragraph{Domain Mixture Ablation.} To study whether all five training domains are necessary and whether our mixture ratios are well-chosen, we compare against two alternatives: a uniform mixture where all five domains contribute equally, and a math-and-code-only mixture, as these two domains have been widely adopted in RL post-training for reasoning models~\citep{deepseek,wang2025nemotron}.
As shown in Table~\ref{tab:domain-mixture-ablation}, our mixture achieves the highest point estimates on all four benchmarks while maintaining competitive output length. The math-and-code-only mixture is competitive on AIME-25 and GPQA but drops on LCB (67.2\% vs. 70.8\%), consistent with a benefit from the remaining three domains. The uniform mixture also has lower point estimates across the benchmarks. Given the reported variance, these comparisons support the selected mixture as a practical configuration rather than establishing that it is uniquely optimal.

\paragraph{Analysis of Reasoning Traces.} To understand \emph{why} {\apriel} produces shorter outputs without sacrificing accuracy, we segment reasoning traces from {\aprielbase} and {\apriel} on AIME 2025 into discrete reasoning steps using GPT-5.4 (see Appendix~\ref{sec:example_trace_analysis} for an example, and Appendix~\ref{sec:trace_analysis_prompts} for the prompts). Figure~\ref{fig:mean_num_steps} shows that the average number of steps per trace is comparable between {\aprielbase} and {\apriel}. However, Figure~\ref{fig:avg_tokens_per_step} demonstrates that {\apriel} uses approximately 35\% fewer tokens per step compared to {\aprielbase} on correct responses. This indicates that RLVR teaches the model to express each reasoning step more concisely.
In Appendix~\ref{sec:appendix-trace-analysis}, we further classify each reasoning step into eight categories, extending the taxonomy of \citet{gandhi2025cognitive}, and show that {\apriel} produces fewer non-productive steps and a higher share of non-linear reasoning behaviors compared to {\aprielbase}.
Our findings suggest that the efficiency gains of {\apriel} do not come from shallower reasoning, but from tighter expression and reduced overhead, consistent with recent work on efficient reasoning in long chain-of-thought models \citep{wang2025thoughts,su2025between}.

\section{Limitations}

Our empirical validation uses one 15B base model and GSPO, so it does not establish cross-architecture, cross-scale, or cross-algorithm generality. The realized-proportion diagnostics verify that adaptive sampling tracks its configured completed-rollout mixture, but we do not include a matched static-sampler run and therefore do not isolate its downstream accuracy effect. Likewise, the LP--DAP comparison isolates the reward-side length-control change under identical adaptive sampling, but we do not provide matched-compute comparisons against GRPO-LEAD, DLER/DA-DLER, DAST, or AdaptThink. Future work should add matched-base, matched-data, and matched-compute comparisons against these methods and distillation-based reasoning baselines such as DeepSeek-R1-Distill~\citep{deepseek}. Sequential-domain training is also left for future work because it changes the setting and introduces ordering and forgetting effects. Finally, AIME 2025 contains only 30 problems; despite repeated decoding runs, small point-estimate gaps on this benchmark should be interpreted cautiously.

\section{Conclusion}

Apriel-Reasoner is a 15B model trained with a reproducible five-domain RL recipe. Adaptive domain sampling maintains the target mixture, while a difficulty-aware length penalty gives harder problems more room to reason. Together, these mechanisms address heterogeneity in rollout dynamics and problem difficulty without adding auxiliary models or separate optimization stages. Its 32K results reflect preserved base-model capability, not RL-induced extrapolation. Across four benchmarks, it improves on Apriel-Base with 30--50\% shorter traces through more concise, not shallower, reasoning. This offers a practical recipe for efficient general-purpose reasoning models.

\section*{Reproducibility Statement}
We are committed to enabling full reproducibility of this work. Hyperparameters, search spaces, and training details are provided in Section~\ref{sec:exp-setup} and Appendix~\ref{sec:training-details}, and key design decisions are justified in their respective sections: choice of base model in Section~\ref{sec:exp-setup}, policy optimization objective in Section~\ref{sec:gspo}, adaptive domain sampling in Section~\ref{sec:adaptive-sampling}, and length penalty in Section~\ref{sec:dap}.

Our RL training pipeline builds on the public PipelineRL implementation.\footnote{\url{https://github.com/ServiceNow/PipelineRL}} All datasets used for training and evaluation are publicly accessible; training data sources are described in Section~\ref{sec:envs} with further details in Appendix~\ref{sec:appendix-envs}, and evaluation benchmarks in Section~\ref{sec:exp-setup}.

Following the transparency principles advocated by the LLM360 initiative~\citep{llm360}, this paper reports the complete domain mixture, selected training configuration, and hyperparameter search space. The reproducibility claim of this paper rests on the documented recipe, public inputs, and the public PipelineRL implementation rather than on availability of the final checkpoint; the post-trained model weights and task-specific training configurations are not part of this release.

\bibliography{colm2026_conference}
\bibliographystyle{colm2026_conference}

\clearpage
\appendix
\raggedbottom

\section{Additional Method Comparisons and Diagnostics}
\label{sec:appendix-comparisons}

\begin{table}[H]
\centering
{
\small
\setlength{\tabcolsep}{3pt}
\begin{tabularx}{\linewidth}{@{}l>{\raggedright\arraybackslash}X>{\raggedright\arraybackslash}X>{\raggedright\arraybackslash}X@{}}
\toprule
\textbf{Method} & \textbf{Extra model/data stage} & \textbf{Optimization change} & \textbf{RLVR integration} \\
\midrule
LP & None & Scalar reward term only & Minimal \\
DAP (ours) & None; reuses group solve rate & Scalar reward only & Minimal \\
GRPO-LEAD & None & Reward and advantage reweighting & Moderate \\
DLER/DA-DLER & Dynamic discard-and-refill sampling & Normalization, clipping, and truncation schedule & Coupled recipe \\
DAST & Budget-estimation pass and preference pairs & SimPO preference optimization & Separate pipeline \\
AdaptThink & Reference-performance estimate and mode sampling & Constrained objective and importance sampling & New generation modes \\
\bottomrule
\end{tabularx}
}
\caption{Comparison of difficulty-aware length-control approaches. ``None'' means no change relative to a standard online RLVR loop with group-based advantages. DAP is designed as a local reward edit; the table does not imply empirical accuracy superiority.}
\label{tab:length-control-comparison}
\end{table}

\begin{table}[H]
\centering
{
\small
\begin{tabular}{lrrr}
\toprule
\textbf{Domain} & \textbf{Target} & \textbf{Realized} & \textbf{Abs. dev.} \\
\midrule
Mathematics & 40.00\% & 40.05\% & 0.05 pp \\
Coding & 25.00\% & 25.01\% & 0.01 pp \\
Logic & 15.00\% & 14.98\% & 0.02 pp \\
Instruction following & 10.00\% & 9.99\% & 0.01 pp \\
Function calling & 10.00\% & 9.96\% & 0.04 pp \\
\midrule
Mean absolute deviation & & & 0.027 pp \\
\bottomrule
\end{tabular}
}
\caption{Configured and realized cumulative proportions of actor-completed rollouts at step 250 of the selected DAP run (683K rollouts), recorded before zero-advantage filtering. Absolute deviations are in percentage points (pp); MAD is computed from unrounded proportions.}
\label{tab:adaptive-sampling-realized}
\end{table}

\begin{figure}[H]
    \centering
    \includegraphics[width=\linewidth]{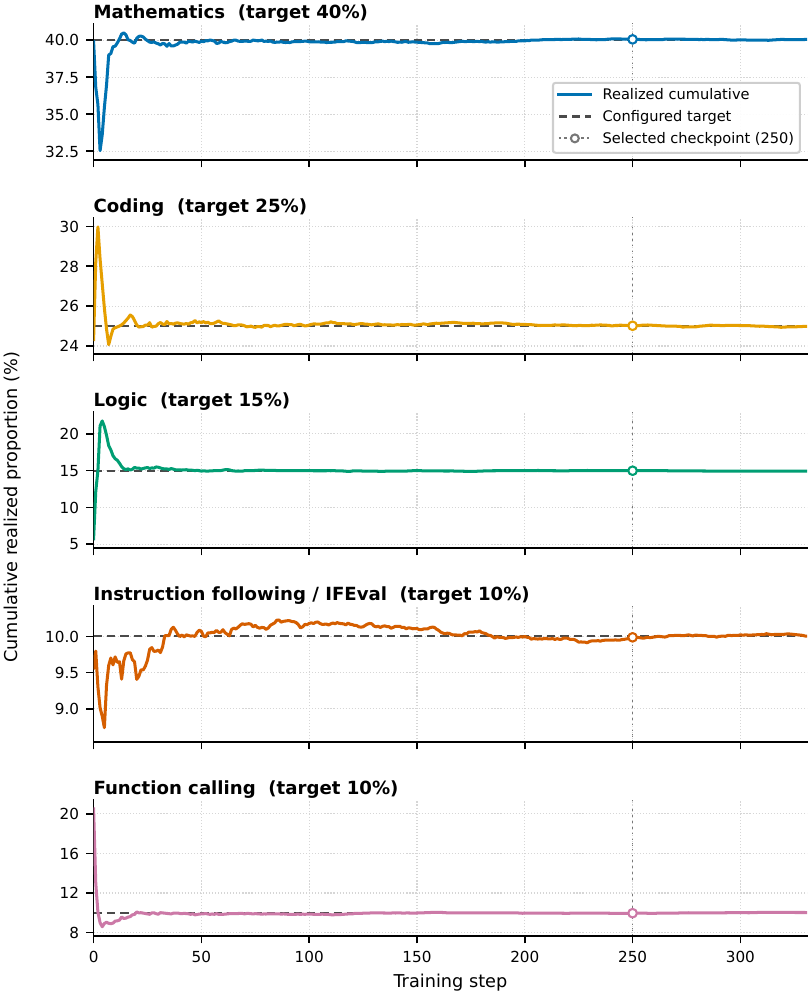}
    \caption{Cumulative realized proportions of completed actor rollouts for the selected DAP run. Solid curves show each domain's cumulative share; dashed horizontal lines show the configured targets; the dotted vertical line and open circles mark the selected step-250 checkpoint. Step 0 is the first non-empty logged window: sampling begins from static weights and may start adapting after 50 completions within that window. Counts are recorded after complete rollout groups return and before zero-advantage filtering. The series ends at the last complete domain-statistics record (step 331; 952,368 rollouts); 608 subsequent completed rollouts without a domain breakdown are excluded. Panels use different vertical scales.}
    \label{fig:adaptive-sampling-trajectory}
\end{figure}

\section{Training Domains and Verifiers}
\label{sec:appendix-envs}

Each training domain is defined by a dataset and a verifiable reward function, which together constitute an \emph{environment}. Below, we describe each domain and its corresponding verifier in more detail.

\paragraph{Mathematics:}
For mathematical reasoning, we use the training data curated by Open-Reasoner-Zero~\citep{openreasonerzero}, totaling approximately 129K math problems spanning a range of difficulty levels, from high-school mathematics to competition-level problems.\footnote{Data accessed from \url{https://github.com/Open-Reasoner-Zero/Open-Reasoner-Zero}.} This includes the original 57K collection, sourced from AIME (up to 2023), MATH~\citep{hendrycks2021math}, NuminaMath~\citep{numina_math}, and Tulu3 MATH~\citep{lambert2024tulu}, as well as an extended 72K subset carefully filtered from OpenR1-Math-220k.\footnote{\url{https://hf.co/datasets/open-r1/OpenR1-Math-220k}} We use a binary reward based on answer correctness. Following standard practice, a rule-based verifier extracts the model’s final numerical or symbolic answer and compares it to the ground truth using exact match.

\paragraph{Code Generation:}
The code generation domain consists of programming problems with test cases used for automatic verification. Our primary source is the TACO dataset~\citep{li2023taco}, which contains roughly 24K competition-style programming problems collected from online judge platforms such as CodeChef, Codeforces, HackerRank, and GeeksforGeeks, as well as prior datasets including APPS, CodeContests, and Description2Code. These problems span a broad range of algorithmic topics typical of contemporary competitive programming.
The dataset provides difficulty annotations, from which we exclude problems labeled `HARD' and `VERY\_HARD' resulting in $\sim$24K training problems.
We use a sandboxed environment in which programs produced by the model during training are executed against provided test cases.\footnote{\url{https://github.com/bytedance/SandboxFusion}}
The verifier returns pass/fail diagnostics, and rewards are assigned using a rule-based scheme; a reward of 1 is given only if all test cases pass, and 0 otherwise.

\paragraph{Instruction-Following:}
In instruction-following, the model must complete a task such as summarization or creative writing while satisfying one or more verifiable output constraints~\citep{zhou2023instruction}.
We train on the IF-RLVR dataset~\citep{pyatkin2025generalizing}, containing $\sim$95K instruction-response pairs (with 550 held out for validation), which was used for training OLMo 3~\citep{olmo3}.
The reward is computed as the proportion of constraints satisfied, ranging from 0 to 1.

\paragraph{Logical Puzzles:}
This domain is adapted from the INTELLECT-3 logic environment~\citep{intellect3}, based on SynLogic~\citep{liu2025synlogic}. The dataset consists of 29 puzzle types and $\sim$12K tasks spanning symbolic and deductive reasoning, constraint satisfaction, and combinatorial search, including Boolean expressions, grid-based puzzles (e.g., Sudoku), arithmetic puzzles (e.g., Game of 24), and interactive logic games (e.g., Minesweeper). Problems are programmatically generated with adjustable difficulty. Rewards are computed using task-specific verifiers from the \texttt{i3-logic} library.\footnote{\url{https://app.primeintellect.ai/dashboard/environments/primeintellect/i3-logic}}

\paragraph{Function Calling:}
The goal of the function calling task is to generate a valid call to an external function or API in response to a user query~\citep{bfcl}. We train and evaluate this domain using all subcategories within the single-turn category of BFCL v4.\footnote{\url{https://gorilla.cs.berkeley.edu/blogs/15_bfcl_v4_web_search.html}}
To compute reward, the predicted function name and arguments are extracted using Python's \texttt{ast} module. Following BFCL's evaluation protocol, a prediction is considered correct if the function name matches exactly and all argument values fall within predefined valid sets, rather than requiring exact string matches.

\section{Training Details}
\label{sec:training-details}

\subsection{Hyperparameters}
\label{sec:hyperparamters}

Table~\ref{tab:gspo-dap-launch-config} lists the full hyperparameters for the training configuration described in Section~\ref{sec:exp-setup}.
In the \textit{Scope} column, \textit{Actor} means rollout generation and reward computation, \textit{Trainer} denotes optimization and model weight updates, and \textit{Both} refers to parameters that directly affect both sides.

\begin{table}[H]
\centering
\begin{tabular}{>{\raggedright\arraybackslash}p{0.32\linewidth}>{\raggedright\arraybackslash}p{0.29\linewidth}cc}
\toprule
\textbf{Parameter} & \textbf{Value} & \textbf{Scope} & \textbf{Status} \\
\midrule
Policy loss & GSPO & Trainer & Fixed \\
FP32 \texttt{lm\_head} & true & Both & Fixed \\
Group Size ($G$) & 8 & Both & Fixed \\
Max output tokens & 16,000 & Actor & Fixed \\
Learning rate & $1.25 \times 10^{-6}$ & Trainer & Swept \\
Clip $(\epsilon_{\text{low}}, \epsilon_{\text{high}})$ & $3\times10^{-3}, 4\times10^{-3}$ & Trainer & Swept \\
DAP $(B,\gamma,\lambda_f)$ & $(2000,1.0,1.0)$ & Actor & Swept \\
Domain rates (M/C/L/I/F) & 0.40/0.25/0.15/0.10/0.10 & Actor & Swept \\
World fractions (Actor/Preprocessor/Trainer) & 4/0/4 & Both & Fixed \\
\bottomrule
\end{tabular}
\caption{Reference {\apriel} launch configuration using GSPO and DAP.}
\label{tab:gspo-dap-launch-config}
\end{table}

\subsection{Hyperparameter Search Space}
\label{sec:appendix-sweeps}

\begin{table}[H]
\centering
{
\small
\begin{tabular}{>{\raggedright\arraybackslash}p{0.28\linewidth}>{\raggedright\arraybackslash}p{0.62\linewidth}}
\toprule
\textbf{Parameter} & \textbf{Candidate values} \\
\midrule
Learning rate & $\{7.5\times10^{-7},\, 1.0\times10^{-6},\, 1.25\times10^{-6},\, 2.5\times10^{-6},\, 5.0\times10^{-6}\}$ \\
GSPO clip $(\epsilon_{\text{low}}, \epsilon_{\text{high}})$ & $\{(3\times10^{-4},\,4\times10^{-4}),\; (3\times10^{-3},\,4\times10^{-3})\}$ \\
Domain rates $(r_i)$ & \begin{tabular}[t]{@{}l@{}}
$(0.20,\,0.20,\,0.20,\,0.20,\,0.20)$ \\
$(0.35,\,0.25,\,0.15,\,0.15,\,0.10)$ \\
$(0.40,\,0.25,\,0.15,\,0.10,\,0.10)$ \\
$(0.30,\,0.30,\,0.15,\,0.15,\,0.10)$ \\
$(0.35,\,0.20,\,0.15,\,0.15,\,0.15)$ \\
$(0.50,\,0.50,\,0.00,\,0.00,\,0.00)$
\end{tabular} \\
DAP $(B,\gamma,\lambda_f)$ & $\{(2000,\,1.0,\,1.0),\; (2000,\,0.5,\,0.5)\}$ \\
Output-token budget $L_{\text{out}}$ & $\{16\text{k},\,20\text{k},\,24\text{k}\}$ \\
\bottomrule
\end{tabular}
}
\caption{Exact candidate sets used in the hyperparameter sweeps reported in this paper. Domain-rate vectors are ordered as (M/C/L/I/F) = (math, coding, logic, ifeval, function calling).}
\label{tab:gspo-dap-sweep-values}
\end{table}

\section{Solve-Rate-Conditioned DAP Analysis}
\label{sec:appendix-dap-conditioned}

We group the 30 AIME 2025 problems by the LP model's solve rate, which provides one shared difficulty partition for comparing LP and DAP. Lengths in \Cref{tab:dap-conditioned-lengths} are mean output tokens per rollout conditional on correctness within each bin. The hardest bin shows that DAP can allocate additional reasoning budget rather than uniformly truncate difficult attempts; the easiest two bins with at least 50\% LP solve rate show shorter correct reasoning under DAP. The $[25\%,50\%)$ and $[50\%,75\%)$ bins contain only one and four problems, respectively, and should not be interpreted independently as stable estimates.

\begin{table}[H]
\centering
{
\small
\setlength{\tabcolsep}{2.6pt}
\begin{tabular}{lrrrrrrrrr}
\toprule
& & \multicolumn{2}{c}{\textbf{Accuracy (\%)}} & \multicolumn{3}{c}{\textbf{Correct length}} & \multicolumn{3}{c}{\textbf{Incorrect length}} \\
\cmidrule(lr){3-4}\cmidrule(lr){5-7}\cmidrule(lr){8-10}
\textbf{LP solve-rate bin} & $n$ & \textbf{LP} & \textbf{DAP} & \textbf{LP} & \textbf{DAP} & $\Delta$ & \textbf{LP} & \textbf{DAP} & $\Delta$ \\
\midrule
$[0,25)$ & 6 & 10 & 22 & 21,766 & 23,259 & +1,493 & 22,810 & 27,590 & +4,780 \\
$[25,50)$ & 1 & 30 & 80 & 5,499 & 6,310 & +810 & 5,355 & 7,148 & +1,793 \\
$[50,75)$ & 4 & 60 & 88 & 15,192 & 12,320 & -2,872 & 18,915 & 23,401 & +4,485 \\
$[75,100]$ & 19 & 96 & 94 & 6,170 & 5,818 & -352 & 14,608 & 11,958 & -2,650 \\
\bottomrule
\end{tabular}
}
\caption{AIME 2025 accuracy and mean output tokens conditioned on correctness within bins defined by LP solve rate. Differences are DAP minus LP.}
\label{tab:dap-conditioned-lengths}
\end{table}

\section{Per-Domain Reward Curves}
\label{sec:appendix-reward-curves}

\begin{figure}[H]
    \centering
    \begin{subfigure}{0.48\textwidth}
        \centering
        \includegraphics[width=\linewidth]{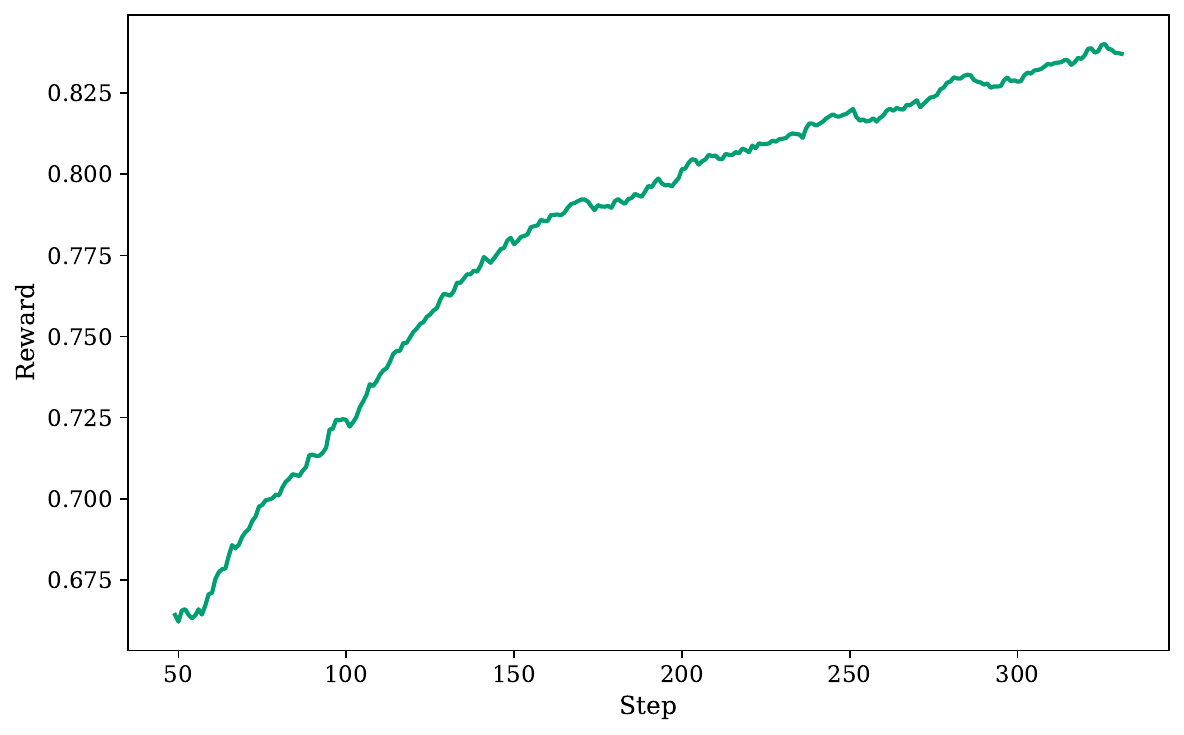}
        \caption{Mathematics}
    \end{subfigure}
    \hfill
    \begin{subfigure}{0.48\textwidth}
        \centering
        \includegraphics[width=\linewidth]{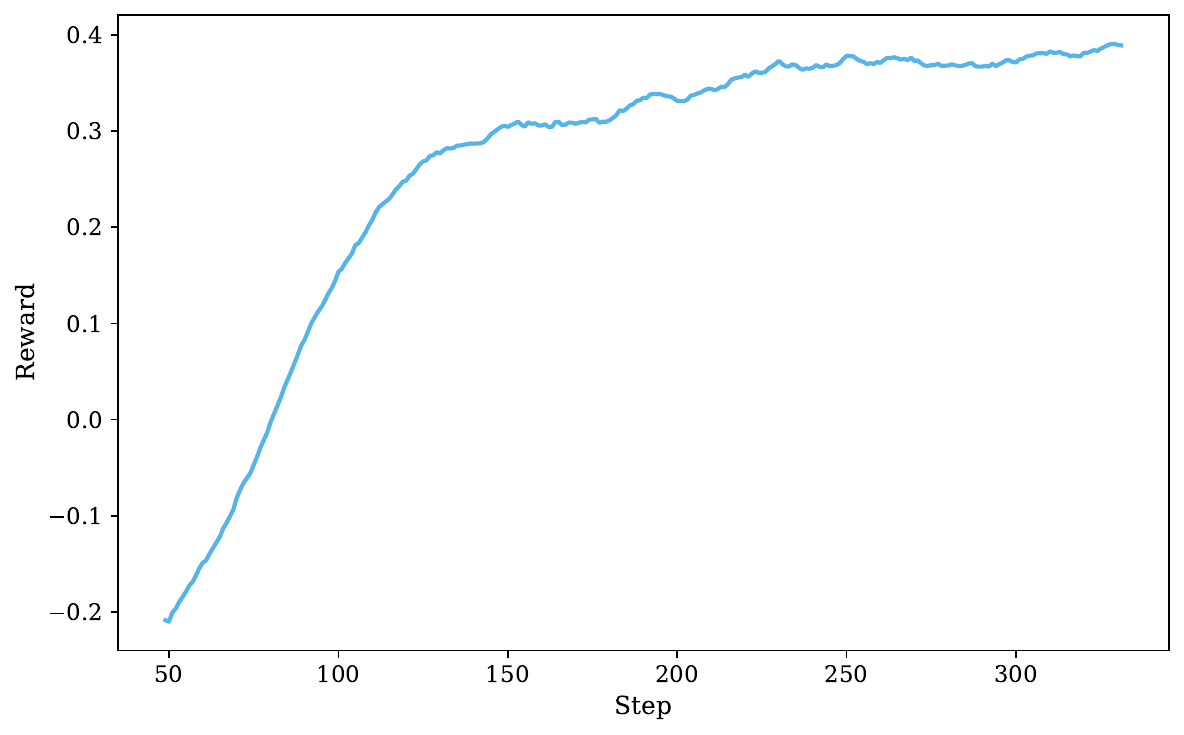}
        \caption{Code Generation}
    \end{subfigure}
    \\[0.5em]
    \begin{subfigure}{0.32\textwidth}
        \centering
        \includegraphics[width=\linewidth]{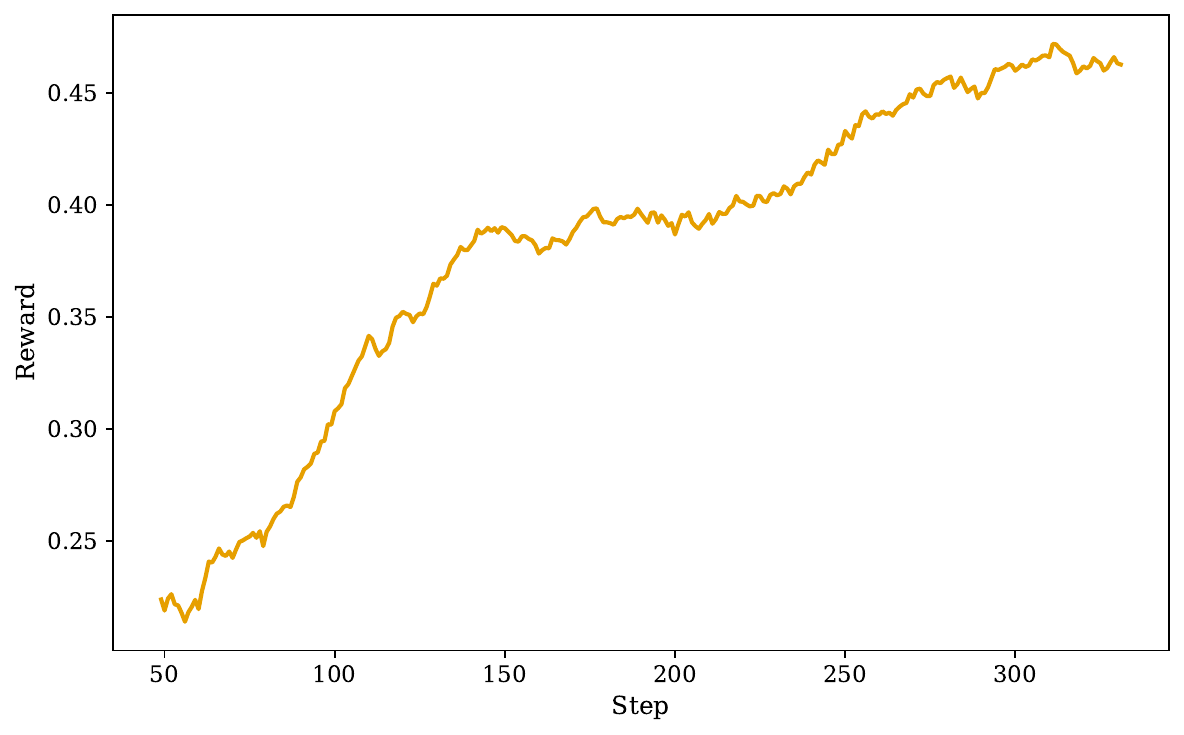}
        \caption{Instruction-Following}
    \end{subfigure}
    \hfill
    \begin{subfigure}{0.32\textwidth}
        \centering
        \includegraphics[width=\linewidth]{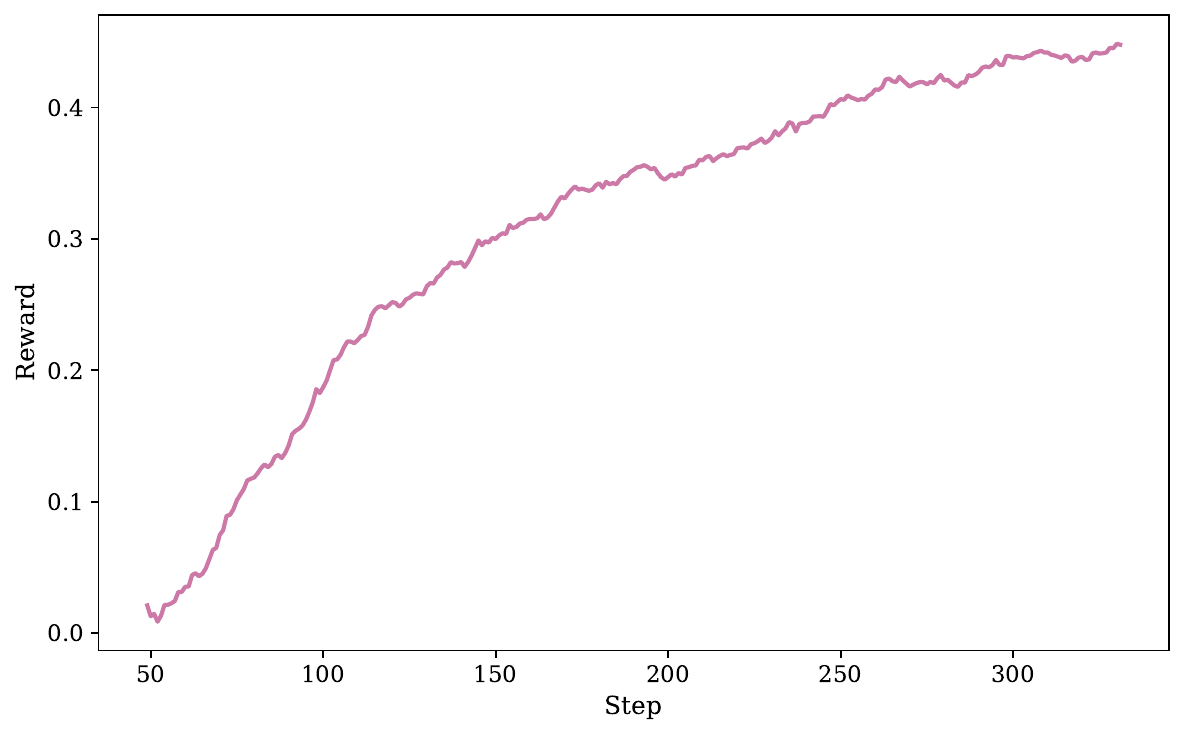}
        \caption{Logical Puzzles}
    \end{subfigure}
    \hfill
    \begin{subfigure}{0.32\textwidth}
        \centering
        \includegraphics[width=\linewidth]{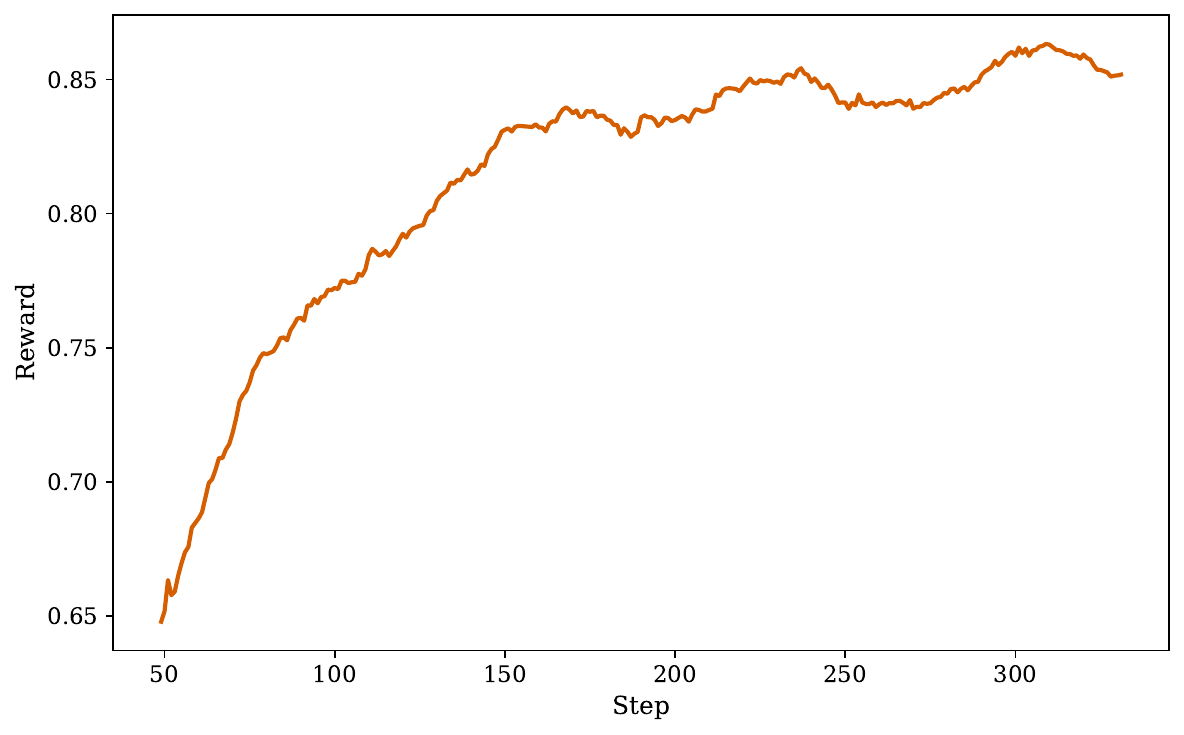}
        \caption{Function Calling}
    \end{subfigure}
    \caption{Mean reward progress during training for each domain.}
    \label{fig:reward_mean}
\end{figure}

\section{Details of Reasoning Trace Analysis}
\label{sec:appendix-trace-analysis}

\begin{figure}[H]
    \centering
    \begin{subfigure}{0.46\linewidth}
        \centering
        \includegraphics[width=\linewidth]{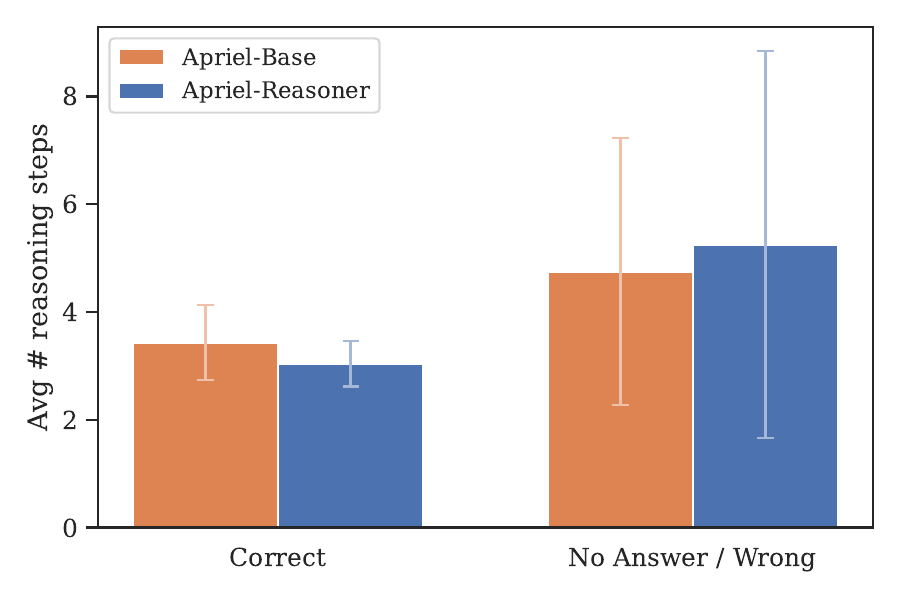}
        \caption{Average reasoning steps per trace}
        \label{fig:mean_num_steps}
    \end{subfigure}
    \hfill
    \begin{subfigure}{0.46\linewidth}
        \centering
        \includegraphics[width=\linewidth]{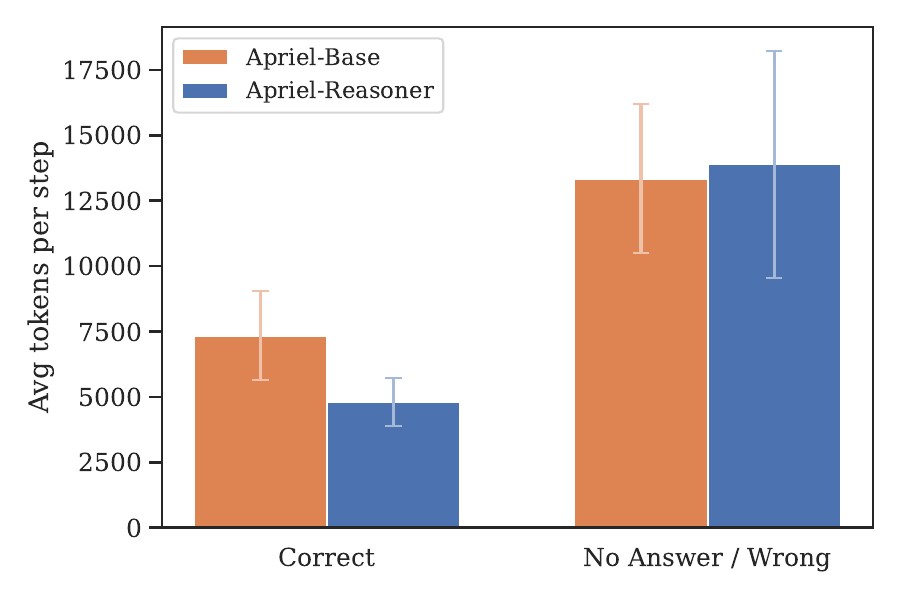}
        \caption{Average tokens per step}
        \label{fig:avg_tokens_per_step}
    \end{subfigure}
    \caption{Average number of reasoning steps per trace (\ref{fig:mean_num_steps}) and average number of tokens per step (\ref{fig:avg_tokens_per_step}) for {\aprielbase} and {\apriel} on AIME 2025. The number of steps is comparable, but {\apriel} expresses each step more concisely.}
    \label{fig:steps_and_tokens}
\end{figure}

Complementary to the reasoning step analysis in Section~\ref{sec:main-results}, we further analyze the structure of each reasoning step into eight categories. Four correspond to the non-linear cognitive behaviors identified by \citet{gandhi2025cognitive}:
\emph{verification}, systematically checking intermediate results against solution criteria to confirm or reject a candidate answer; \emph{backtracking}, explicitly abandoning a failing line of reasoning and pivoting to an alternative approach after recognizing an error or dead end; \emph{subgoal setting}, decomposing a complex problem into smaller, manageable intermediate targets before solving each one; and \emph{backward chaining}, reasoning from the desired outcome back toward the inputs, asking what conditions must hold for the goal to be satisfied.
These four behaviors represent non-linear reasoning patterns that go beyond the typical linear, monotonic chain-of-thought commonly observed in LLMs.

To fully account for all content in a reasoning trace, we introduce four additional categories complementary to the above:
\emph{problem setup} for steps that translate the problem into a workable form, such as interpreting the problem statement, converting representations, or defining variables before solving begins; \emph{forward reasoning} as the default category for standard linear deduction; \emph{final answer}, for the steps where the model produces its final answer; and \emph{non-productive} steps, which capture content that restates a previously established result without adding new computation or verification, including repeated statements of the answer, re-derivation of a conclusion already reached earlier in the trace, and filler meta-commentary that adds no new information.
This last category relates to the \emph{overthinking} phenomenon documented by \citet{chen2025not}, where models waste tokens revisiting already-established results.

\begin{figure}[H]
    \centering
    \includegraphics[width=\linewidth]{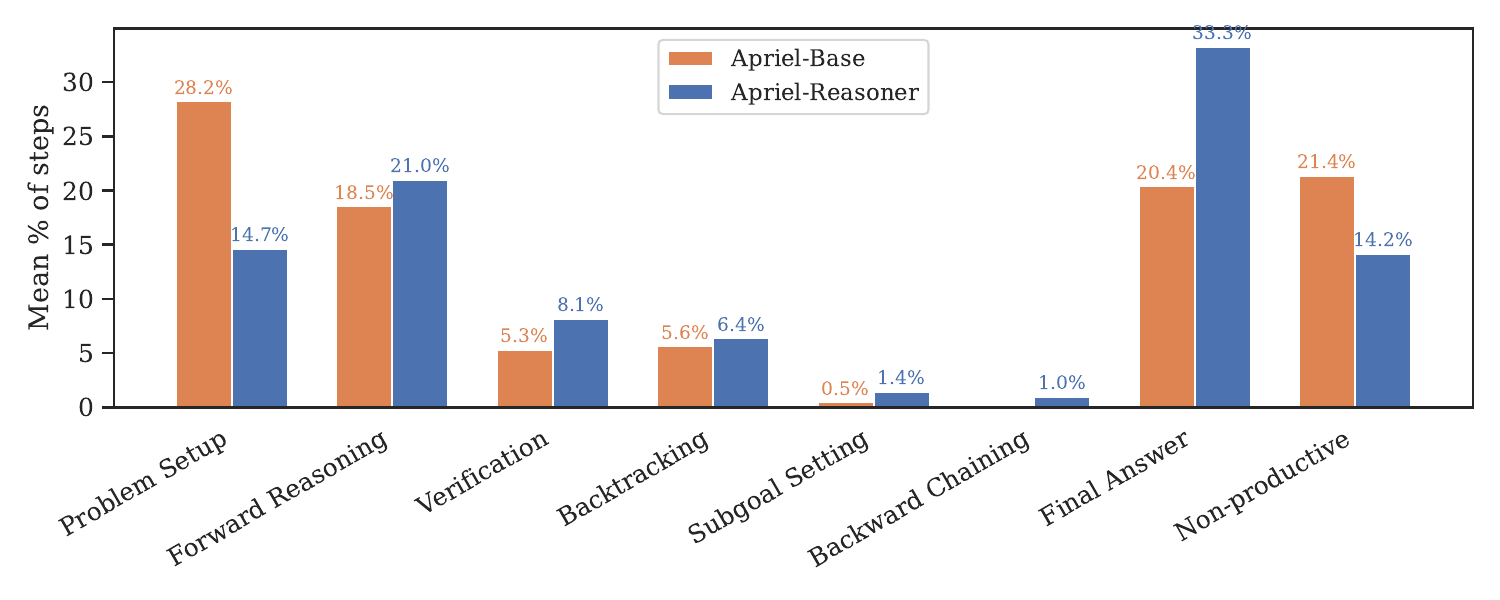}
    \caption{Step-type distribution for {\aprielbase} and {\apriel} on AIME 2025. {\apriel} produces fewer non-productive steps and more verification, backtracking, subgoal-setting, and backward-chaining steps.}
    \label{fig:step_types}
\end{figure}

Using the same segmentation described in Section~\ref{sec:main-results}, we classify each step into one of the eight categories above using GPT-5.4. The full step-type distribution is shown in Figure~\ref{fig:step_types}.

\paragraph{Fewer non-productive steps and more non-linear reasoning.}
The most notable difference is in \emph{non-productive} steps, which drop from 21.4\% in {\aprielbase} to 14.2\% in {\apriel}, meaning the model spends fewer steps on content that does not advance the solution. Meanwhile, the combined share of non-linear reasoning behaviors (verification, backtracking, subgoal setting, and backward chaining) increases from approximately 11\% to 17\%. Verification rises from 5.3\% to 8.1\%, backtracking from 5.6\% to 6.4\%, subgoal setting from 0.5\% to 1.4\%, and backward chaining from a value that rounds to 0\% to 1.0\%. Backward chaining therefore remains rare, and we do not draw a standalone conclusion from this shift. When it appears, the trace begins from a target quantity or final constraint and works backward through conditions that must hold, rather than enumerating computations only in forward order. These descriptive results suggest that RL training changes the composition of reasoning steps, but they do not establish a causal mechanism.

\subsection{Example of Reasoning Trace Analysis}
\label{sec:example_trace_analysis}

\begin{figure}[H]
    \centering
    \includegraphics[width=\linewidth]{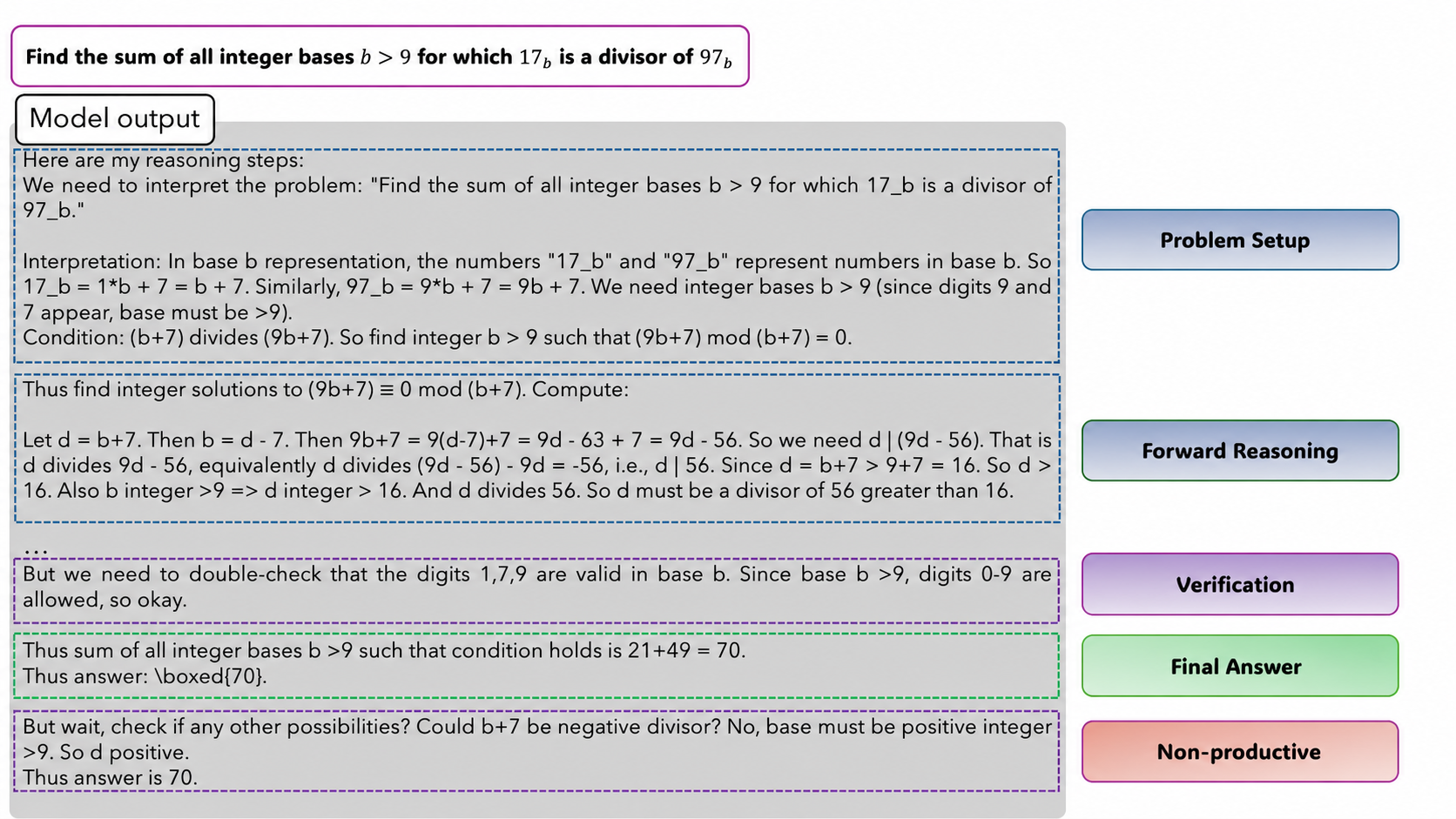}
    \caption{An example reasoning trace from {\aprielbase} on a problem from AIME 2025, first split into reasoning steps and then classified into the eight categories defined in Section~\ref{sec:appendix-trace-analysis}. The ellipsis marks an omitted intermediate forward-reasoning segment. The model output is reformatted for readability.}
    \label{fig:trace_analysis_example}
\end{figure}

\subsection{Prompts}
\label{sec:trace_analysis_prompts}

\includeprompt{Extracting Steps from Reasoning Traces}{prompts/step_extraction.txt}

\includeprompt{Classifying Reasoning Steps}{prompts/classify_steps.txt}

\end{document}